\documentclass[10pt,a4paper]{article}

\usepackage[top=2.2cm,bottom=2.5cm,left=1.8cm,right=1.8cm,columnsep=0.65cm]{geometry}
\usepackage[utf8]{inputenc}
\usepackage[T1]{fontenc}
\usepackage{lmodern}
\usepackage{amsmath,amssymb}
\usepackage{booktabs}
\usepackage{graphicx}
\graphicspath{{figures/}}
\usepackage{xcolor}
\usepackage{hyperref}
\usepackage[numbers,sort&compress]{natbib}
\usepackage{float}
\usepackage{stfloats}
\usepackage{array}
\usepackage{tabularx}

\definecolor{navy}{HTML}{2E4057}
\definecolor{teal}{HTML}{048A81}
\definecolor{coral}{HTML}{E07A5F}

\hypersetup{
  colorlinks=true,
  linkcolor=navy,
  citecolor=teal,
  urlcolor=coral,
  pdftitle={Auditing Construct Overlap in Explainable Machine Learning: Evidence from Burnout--Depression Prediction Across Student Cohorts},
  pdfauthor={Alireza Dehghan, Negin Ashrafi},
}

\makeatletter
\renewcommand\section{\@startsection{section}{1}{\z@}%
  {-2.2ex \@plus -0.5ex \@minus -.2ex}{0.9ex \@plus .1ex}%
  {\normalfont\large\bfseries}}
\renewcommand\subsection{\@startsection{subsection}{2}{\z@}%
  {-1.8ex \@plus -0.4ex \@minus -.2ex}{0.5ex \@plus .1ex}%
  {\normalfont\normalsize\bfseries}}
\renewcommand\subsubsection{\@startsection{subsubsection}{3}{\z@}%
  {-1.4ex \@plus -0.3ex \@minus -.2ex}{0.3ex \@plus .1ex}%
  {\normalfont\normalsize\itshape}}
\long\def\@makecaption#1#2{%
  \vskip\abovecaptionskip
  \sbox\@tempboxa{{\small\textbf{#1.} #2}}%
  \ifdim \wd\@tempboxa >\hsize
    {\small\textbf{#1.} #2\par}%
  \else
    \global\@minipagefalse
    \hb@xt@\hsize{\hfil\box\@tempboxa\hfil}%
  \fi
  \vskip\belowcaptionskip}
\def\ps@myheadings{%
  \def\@oddhead{\hfil{\small\itshape Dehghan \& Ashrafi:
    Auditing Construct Overlap in Explainable ML}\hfil}%
  \let\@evenhead\@oddhead
  \def\@oddfoot{\hfil\small\thepage\hfil}%
  \let\@evenfoot\@oddfoot}
\def\ps@firstpagestyle{%
  \def\@oddhead{\hfill{\footnotesize Submitted Manuscript}}%
  \let\@evenhead\@oddhead
  \def\@oddfoot{\hfil\small\thepage\hfil}%
  \let\@evenfoot\@oddfoot}
\makeatother

\begin{document}

\twocolumn[{%
\begin{center}
{\LARGE\bfseries Auditing Construct Overlap in Explainable Machine Learning:
Evidence from Burnout-Depression Prediction
Across Student Cohorts}

\vspace{10pt}

{\normalsize
\textbf{Alireza Dehghan}\textsuperscript{1}\quad
\textbf{Negin Ashrafi}\textsuperscript{2,*}
}

\vspace{4pt}

{\small
\textsuperscript{1}Sharif University of Technology, Tehran, Iran\\
\textsuperscript{2}University of Southern California, California, USA\\
\textsuperscript{*}Corresponding author:
\href{mailto:ashrafin@usc.edu}{ashrafin@usc.edu}
}

\vspace{10pt}

\begin{minipage}{0.93\textwidth}
\small
\textbf{Abstract.}
Explainable machine learning (XML) pipelines applied to composite mental
health outcomes can produce apparently-robust, cross-population-stable risk
hierarchies that are largely artefacts of how the outcome was constructed.
We demonstrate this using an ElasticNet pipeline applied to 886 medical
students at the University of Lausanne (primary cohort, 2022),
validated across 2{,}580 longitudinal observations at three time points
and 701 non-medical students from eight faculties; all three datasets
share identical instruments.

The pipeline produces a hierarchy in which trait anxiety and health
satisfaction dominate wherever the outcome is measured, with
Kendall $\tau = 1.0$ for the top-two positions across all five evaluation
sets and consistent transfer performance (R\textsuperscript{2}: 0.41--0.49).
Two residualization experiments, which isolate shared variance between correlated variables via regression, reveal the mechanism: when trait anxiety
(STAI-T) is residualized against the co-included depression subscale
(CES-D, $r = 0.72$), model R\textsuperscript{2} drops from 0.41 to 0.16
and STAI-T falls from rank~1 to rank~6; when burnout subscales are
residualized against CES-D, R\textsuperscript{2} collapses to 0.016.
Prediction intervals average 35.4 units on a 0--100 scale (2.4 outcome
standard deviations), independently ruling out individual-level deployment.

The residualization protocol is the paper's transferable contribution:
any XAI study combining correlated predictor and outcome constructs should
apply this check before interpreting apparent stability as a finding.
\end{minipage}

\vspace{8pt}


\vspace{12pt}
\end{center}
}]
\thispagestyle{firstpagestyle}
\pagestyle{myheadings}

%% =============================================================================
\section{Introduction}
\label{sec:intro}
%% =============================================================================

Burnout and depression among medical students are prevalent and consequential:
reported burnout rates consistently exceed those of the general population,
and untreated distress impairs clinical empathy and patient care quality
\cite{AlAlawi2019,Pokhrel2020,Morcos2022}. The application of machine
learning to self-report survey data offers a route to automated risk
stratification; paired with SHAP explanations, such pipelines appear to
identify stable, cross-population risk hierarchies, a consistent rank-ordering of predictors by importance, in which trait anxiety and health satisfaction dominate
\cite{Bozorgmehr2023,Rashid2024}.

The objective of this study is not to identify definitive risk factors, but
to test whether the stability produced by such a pipeline is real.
Specifically: trait anxiety (STAI-T) and the CES-D depression scale are
highly correlated ($r = 0.72$ in this sample). The CES-D is embedded inside
the composite burnout-depression outcome. When a predictor and an outcome
component share a latent construct, the predictor's explanatory dominance
may reflect psychometric overlap rather than a genuine risk relationship.
We call this construct overlap, and we introduce a residualization protocol
to quantify how much of the apparent hierarchy survives once the overlap is
removed.

The principal contributions are threefold. First, we validate a
\textit{residualization protocol for detecting construct overlap artefacts}:
removing the anxiety--depression psychometric channel collapses
R\textsuperscript{2} from 0.41 to 0.016 (burnout-specific outcome) and
STAI-T from rank~1 to rank~6. Second, we demonstrate that
\textit{apparent cross-cohort stability is not evidence of a portable risk
structure} when the signal is anchored to a fixed psychometric correlation:
the model transfers with R\textsuperscript{2} = 0.41--0.49 across five
evaluation sets because the STAI-T/CES-D bivariate relationship is constant,
not because the pipeline learned a generalisable risk structure. Third, we
provide \textit{deployment-relevant uncertainty quantification}: conformal
prediction intervals average 35.4 BDCI units (2.4 outcome standard
deviations), establishing that individual-level clinical prediction is not
actionable with current feature sets.

The remainder of this paper is organised as follows.
Section~\ref{sec:lit} briefly reviews related work.
Section~\ref{sec:method} describes the methodology.
Section~\ref{sec:results} presents results in four subsections structured
around the central audit.
Section~\ref{sec:discussion} discusses implications.
Section~\ref{sec:conclusion} concludes.

%% =============================================================================
\section{Related Work}
\label{sec:lit}
%% =============================================================================

A growing body of work applies machine learning and XAI to mental health
outcomes in student populations. \citet{Martinez2021} achieved high
classification accuracy for teacher burnout using ANN. \citet{Bozorgmehr2023}
applied XGBoost with SHAP to a large German adult sample and identified
health satisfaction as a dominant protective factor, consistent with our
findings on a different instrument. \citet{Rashid2024} applied ensemble and
deep learning to the same 886-student Lausanne cohort as this paper, achieving
moderate classification accuracy. \citet{Benito2024} and \citet{Nnadi2026}
combined SHAP with clustering and hierarchical explanation frameworks. In all
cases, the feature hierarchy is treated as a finding; no prior work has tested
whether the hierarchy survives a residualization check for construct overlap.

Table~\ref{tab:litreview} summarises reviewed studies. The methodological
gap is clear: regression-based continuous framing and construct-overlap
auditing are both absent from the existing literature.

\begin{table*}[ht!]
\centering
\caption{Summary of reviewed studies on machine learning for mental health
  estimation in student and clinical populations.}
\label{tab:litreview}
\renewcommand{\arraystretch}{1.15}
\footnotesize
\begin{tabularx}{\textwidth}{@{}clXXc@{}}
\toprule
\textbf{No.} & \textbf{Authors (Year)} & \textbf{Model / Approach}
  & \textbf{Explainability} & \textbf{N} \\
\midrule
1  & Al-Alawi et al.\ (2019)           & Logistic Regression             & None                   & 662    \\
2  & Wang et al.\ (2019)               & Statistical Analysis             & None                   & 1,271  \\
3  & Cheng et al.\ (2020)              & Linear Regression                & None                   & 1,722  \\
4  & Pokhrel et al.\ (2020)            & Logistic Regression              & None                   & 651    \\
5  & Ilic et al.\ (2021)               & Logistic Regression              & None                   & 760    \\
6  & Mart\'inez-Ram\'on et al.\ (2021) & ANN                              & None                   & 419    \\
7  & Lou et al.\ (2022)                & Penalised LR, SVM, RF            & None                   & 88     \\
8  & Carrard et al.\ (2022)            & Logistic Regression              & None                   & 886    \\
9  & Ghosh et al.\ (2023)              & SVM                              & None                   & 1,182  \\
10 & Alam \& Alam (2023)               & Neural Network                   & Pre-hoc                & 28     \\
11 & Bozorgmehr \& Weltermann (2023)   & XGBoost                          & SHAP                   & 5,801  \\
12 & Benito et al.\ (2024)             & RF, SVM, MLP                     & SHAP, UMAP             & 9,291  \\
13 & Khan et al.\ (2024)               & NLP + Ensemble                   & Feature Importance     & 19,000 \\
14 & Rashid et al.\ (2024)             & RF, NB, DNN                      & None                   & 886    \\
15 & Li et al.\ (2025)                 & Regression + Classification      & SHAP                   & 2,024  \\
16 & Nnadi et al.\ (2026)              & RF, SVM                          & SHAP, H-LIME, DiCE     & 27,901 \\
\midrule
*  & This study  & ElasticNet (primary); OLS, Ridge, Lasso, SVR
                 & SHAP; residualization audit & 886 / 2,580 / 701 \\
\bottomrule
\end{tabularx}
\end{table*}

%% =============================================================================
\section{Methodology}
\label{sec:method}
%% =============================================================================

\subsection{Data and Cohorts}
\label{sec:data}

Table~\ref{tab:cohorts} describes the three datasets. The primary cohort
($n = 886$ medical students, University of Lausanne, 2022) was collected by
the Faculty of Biology and Medicine \cite{Carrard2022}. It contains no
missing values. Predictors include ten demographic and psychological survey
items (age, academic year, health satisfaction, psychotherapy history,
partnership status, employment status, weekly study hours, mother tongue, sex,
and trait anxiety from the State-Trait Anxiety Inventory; STAI-T). Outcome
components are the three Maslach Burnout Inventory--Student Survey (MBI-SS)
subscales and the Center for Epidemiological Studies Depression Scale (CES-D).

\begin{table*}[ht!]
\centering
\caption{Cohort overview. All datasets use identical instruments (CES-D,
  STAI-T, MBI-SS). 2022 thresholds are used for BDCI construction in all cohorts.}
\label{tab:cohorts}
\renewcommand{\arraystretch}{1.15}
\small
\begin{tabularx}{\textwidth}{@{}lXccc@{}}
\toprule
\textbf{Cohort} & \textbf{Description} & \textbf{$n$} & \textbf{Design} & \textbf{Source} \\
\midrule
Primary (2022)   & Medical students, Univ.\ Lausanne     & 886   & Cross-sectional & \cite{Carrard2022} \\
Temporal (2021--22) & Medical, 3 waves (Mar.~2021, Nov.~2021, Nov.~2022) & 2,580 & Longitudinal & \cite{Carrard2024} \\
Non-medical (NMS) & 8 non-medical faculties, same institution & 701 & Cross-sectional & \cite{Carrard2025} \\
\bottomrule
\end{tabularx}
\end{table*}

The temporal and non-medical cohorts are openly available under CC-BY~4.0
(Zenodo: 10.5281/zenodo.8405764 and 10.5281/zenodo.15149289 respectively).

\subsection{BDCI Construction}
\label{sec:bdci}

The Burnout--Depression Composite Index (BDCI) integrates the three MBI-SS
subscales and CES-D into a single severity score on $[0, 100]$.
Each subscale is normalised to its theoretical maximum (reversing the
Academic Efficacy subscale); any component exceeding its empirical
75th-percentile threshold (derived from the 2022 training split) is doubled.
The raw sum is then scaled by $100/(4+k)$, where $k$ is the count of
elevated components. The result is a continuous index where higher values
indicate greater joint burnout-depression severity. Formally:
\begin{equation}
  \mathrm{BDCI} = \frac{100}{4+k}\sum_{j} s_j^*,
  \quad s_j^* = \begin{cases} 2s_j & s_j > \tau_j \\ s_j & \text{otherwise}\end{cases}.
\end{equation}
The 75th-percentile thresholding and doubling scheme is intended to give greater analytic weight to clinically elevated symptom levels, a severity-weighting logic common in composite mental-health indices. We treat this as one reasonable analytic choice rather than a validated clinical rule: the threshold is sample-derived (2022 training split) rather than tied to an external clinical cutoff (e.g., CES-D $\geq$ 16), which is a limitation we return to in Section 5.1. Sensitivity to the weighting choice itself is assessed directly via BDCI-A (Table 5), an equal-weight variant that reproduces the same qualitative pattern, indicating the central finding does not depend on this specific scheme. The BDCI is a study-specific analytical composite; it has not been
independently validated as a clinical instrument.

\begin{figure*}[ht!]
  \centering
  \includegraphics[width=0.72\textwidth]{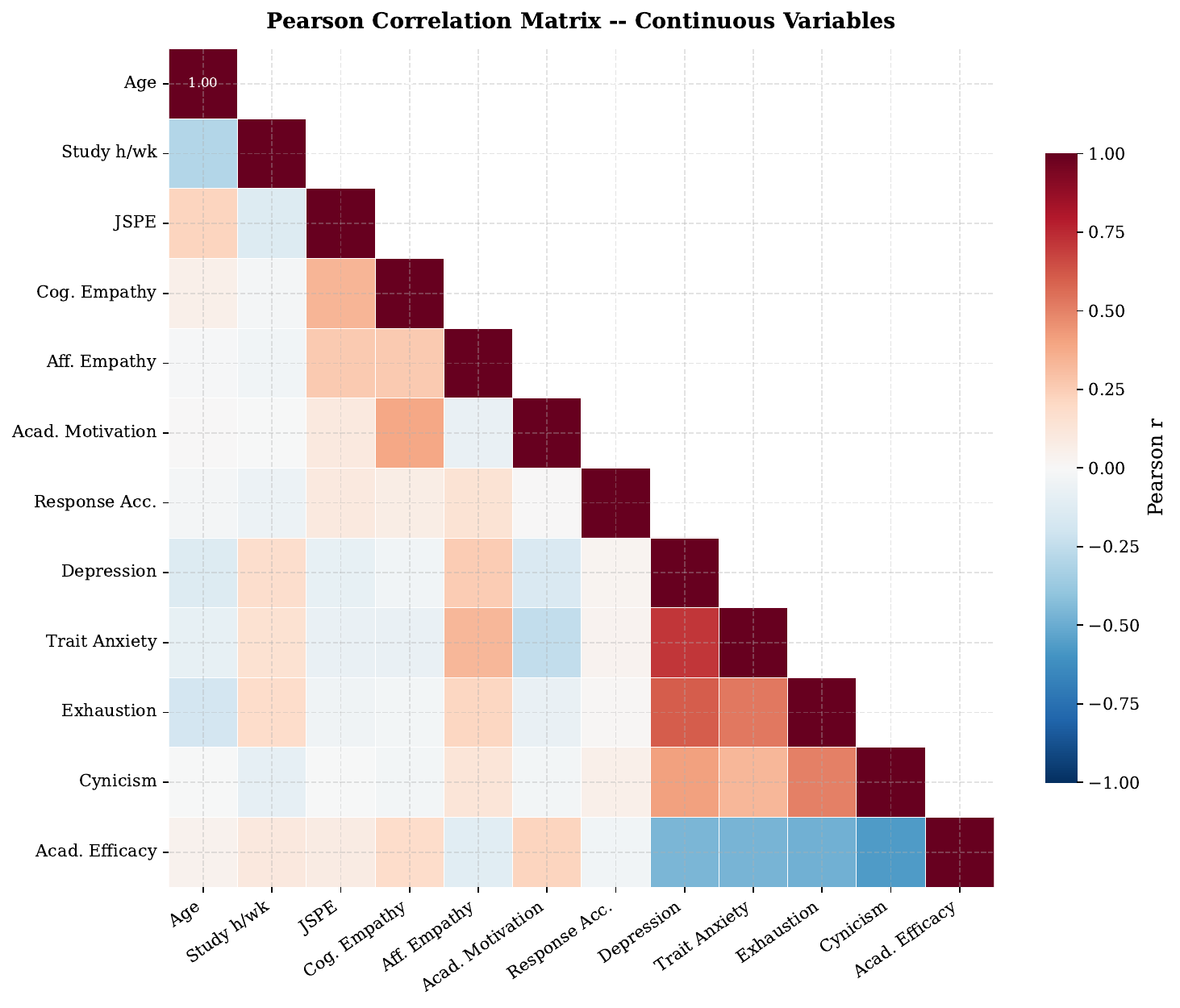}
  \caption{Pearson correlation matrix for the twelve continuous variables.
    The strong anxiety--depression association ($r = 0.72$, STAI-T and CES-D)
    is the structural source of construct overlap: CES-D is an outcome
    component while STAI-T is the dominant predictor.}
  \label{fig:corr_heatmap}
\end{figure*}

The critical structural feature of the BDCI is that CES-D is one of its four
components, while STAI-T (trait anxiety) is one of the predictors.
Figure~\ref{fig:corr_heatmap} shows the Pearson correlation matrix: STAI-T
and CES-D share $r = 0.72$ ($p < 0.001$), the strongest pairwise correlation
in the dataset. This overlap motivates the residualization audit in Section~\ref{sec:audit}.

\subsection{Analysis Pipeline}
\label{sec:pipeline}

The dataset was partitioned 80/20 into training ($n_{\mathrm{train}} = 708$)
and test ($n_{\mathrm{test}} = 178$) sets using stratified random sampling
on BDCI quartile (random seed 42 throughout). All predictors were
$z$-score normalised using training-split statistics only. ElasticNet
(L1+L2 regularisation) was selected as the primary model after 5-fold
cross-validated comparison of six candidate regressors (OLS, Ridge, Lasso,
ElasticNet, SVR, ANN); hyperparameters were selected by grid search
($\alpha \in [10^{-4}, 10^{1}]$, mixing ratio $\rho \in \{0.1, 0.3, 0.5,
0.7, 0.9\}$; optimal: $\alpha = 0.0316$, $\rho = 0.7$). SHAP values were
computed using \texttt{LinearExplainer} \cite{Saeed2023}.

For cross-cohort validation, a 7-feature reduced ElasticNet was
retrained on the 2022 training split using only the predictors available
in all three cohorts (age, academic year, health satisfaction, psychotherapy
history, partnership status, STAI-T, sex), then applied without retraining
to the temporal and non-medical cohorts. The 2022 BDCI thresholds
$\tau_j$ were used throughout to prevent leakage.

Ninety-five percent bootstrap confidence intervals (1,000 resamples) were
computed for all test-set metrics.

\subsection{Residualization Protocol}
\label{sec:resid_method}

Two orthogonality experiments were conducted to isolate the anxiety--depression
channel. \textit{Experiment~1 (predictor residualization)}: STAI-T was
replaced by the residuals of regressing STAI-T on CES-D (fit on training
set, applied to test set). The residual captures trait anxiety variance
orthogonal to concurrent depression. \textit{Experiment~2 (outcome
residualization)}: each MBI-SS subscale was individually residualized
against CES-D (OLS, training set), then re-normalised and assembled into
a depression-free composite (BDCI-C). In both cases, ElasticNet was refit
from scratch on the modified data using the same hyperparameter grid and
split. We note that BDCI-C is a statistical construct created solely to isolate depression-independent variance in the burnout subscales; residualizing against CES-D does not by itself produce a validated 'pure burnout' measure. A near-zero R² for BDCI-C indicates that this feature set cannot predict the depression-independent portion of the composite, not that a clinically meaningful burnout-specific construct doesn't exist. Results are compared with the primary BDCI and a CES-D-excluded
composite (BDCI-B, MBI-SS only) in Table~\ref{tab:audit}.

\subsection{Conformal Prediction}
\label{sec:conformal}

Point estimates are complemented by split conformal prediction intervals
\cite{Vovk2005}. The test set is split 50/50: the calibration half
($n_{\mathrm{cal}} = 89$) is used to compute non-conformity scores
$s_i = |y_i - \hat{y}_i|$; the evaluation half ($n_{\mathrm{eval}} = 89$)
receives intervals $[\hat{y}_j \pm \hat{q}]$, where $\hat{q}$ is the
$(1-\alpha)(1+1/n_{\mathrm{cal}})$ quantile at $\alpha = 0.10$.
This procedure provides marginal coverage $\geq 90\%$ without distributional
assumptions, with the caveat that small calibration sets may produce slight
deviations from the nominal level.

%% =============================================================================
\section{Results}
\label{sec:results}
%% =============================================================================

\subsection{A Standard Pipeline Produces an Apparent Hierarchy}
\label{sec:baseline}

Table~\ref{tab:results} presents test-set performance. All four linear and
regularised models achieve similar accuracy (R\textsuperscript{2}:
0.400--0.414); SVR performs somewhat worse. Penalisation offers no
meaningful gain over OLS: the BDCI--predictor relationship is approximately
linear and additive. A Lasso fit with L1 penalty automatically zeros five
of twelve features, retaining a sparse set of seven predictors.

\begin{table}[H]
\centering
\caption{Test-set performance (95\% bootstrap CI, 1,000 resamples). All four
  linear variants are statistically indistinguishable.}
\label{tab:results}
\renewcommand{\arraystretch}{1.2}
\small
\begin{tabular}{@{}lcc@{}}
\toprule
\textbf{Model} & \textbf{R\textsuperscript{2} [95\% CI]}
               & \textbf{MAE [95\% CI]} \\
\midrule
OLS        & 0.400 [0.257, 0.509] & 9.38 [8.36, 10.52] \\
Ridge      & 0.401 [0.262, 0.506] & 9.38 [8.36, 10.52] \\
Lasso      & 0.413 [0.281, 0.515] & 9.28 [8.27, 10.38] \\
ElasticNet & 0.414 [0.285, 0.514] & 9.26 [8.27, 10.35] \\
SVR        & 0.388 [0.222, 0.506] & 9.43 [8.34, 10.52] \\
ANN        & 0.375 [0.242, 0.481] & 9.59 [8.49, 10.73] \\
\bottomrule
\end{tabular}
\end{table}

\begin{figure*}[ht!]
  \centering
  \includegraphics[width=0.7\textwidth]{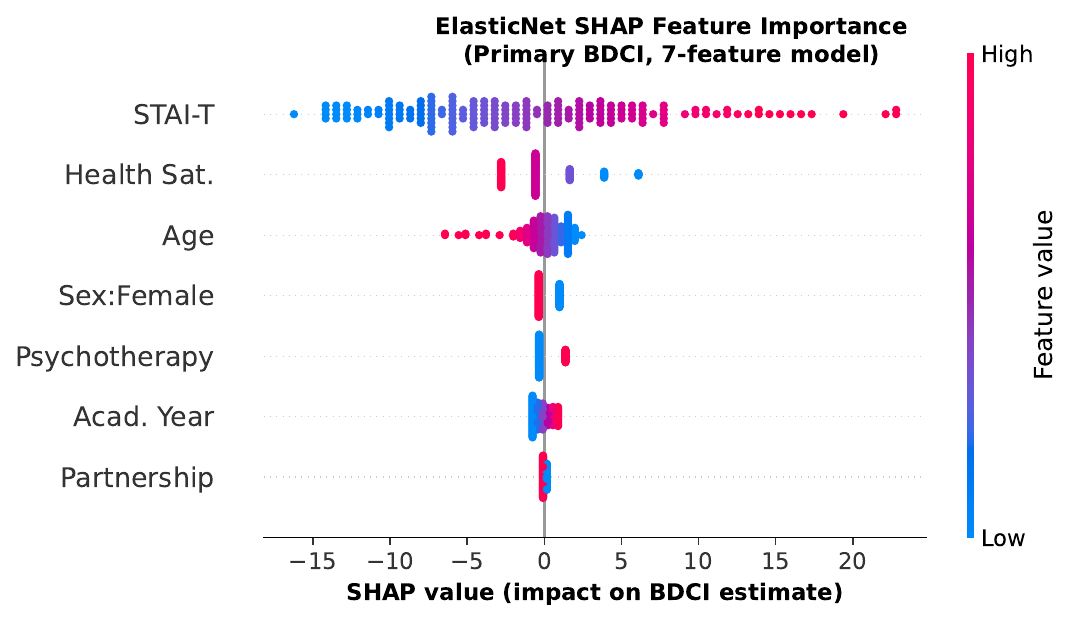}
  \caption{ElasticNet SHAP beeswarm (primary BDCI, 7-feature model). Each
    point is one test-set observation; colour encodes the raw feature value
    (red = high, blue = low). Trait anxiety (STAI-T) dominates by
    approximately 10-fold over health satisfaction.}
  \label{fig:shap_bee}
\end{figure*}

Figure~\ref{fig:shap_bee} shows the ElasticNet SHAP beeswarm. Trait anxiety
produces the widest spread of SHAP values by a wide margin: its mean
$|\text{SHAP}|$ exceeds health satisfaction (ranked second) by approximately
10-fold. This apparent hierarchy is consistent across the cross-population
transfer cohorts (Section~\ref{sec:transfer}): STAI-T ranks first in
every wave and in the non-medical student sample.

\subsection{The Hierarchy Appears Stable Across Cohorts}
\label{sec:transfer}

\begin{figure*}[ht!]
  \centering
  \includegraphics[width=\textwidth]{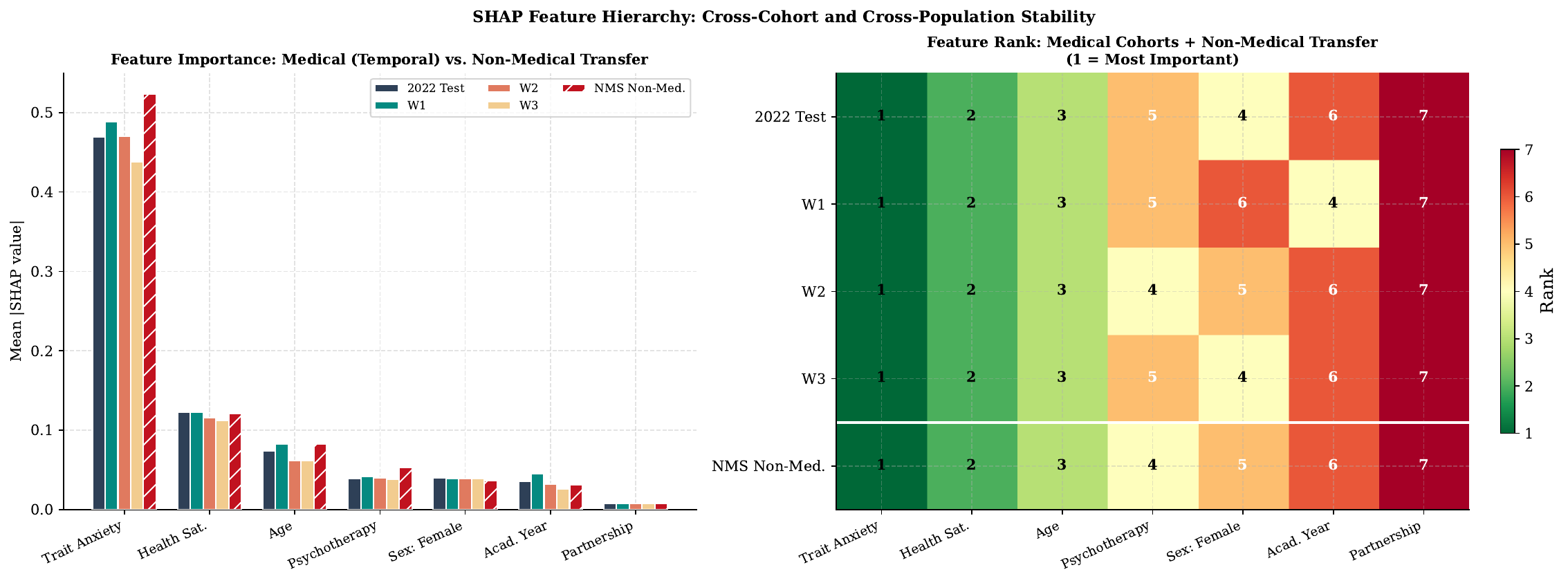}
  \caption{SHAP feature importance across all five evaluation cohorts.
    \textit{Left}: mean $|\text{SHAP}|$ by feature and cohort. \textit{Right}:
    Feature importance rank heatmap (1 = most important; the white line
    separates medical from non-medical cohorts). Trait anxiety holds rank~1
    and health satisfaction holds rank~2 in all five cohorts.}
  \label{fig:crosspop_shap}
\end{figure*}

Table~\ref{tab:transfer} shows transfer performance. The 7-feature reduced
ElasticNet, trained exclusively on 2022 data, achieves
R\textsuperscript{2} of 0.443, 0.487, and 0.388 on the three longitudinal
medical waves, and 0.493 on 701 non-medical students from eight faculties,
the highest of any evaluation set. Figure~\ref{fig:crosspop_shap}
shows SHAP feature importance across all five cohorts. Trait anxiety holds
rank~1 and health satisfaction holds rank~2 in every cohort; the full rank
ordering is Trait Anxiety $>$ Health Satisfaction $>$ Age $>$ Psychotherapy
$>$ Sex:Female $>$ Partnership $>$ Academic Year in all five evaluation sets.
Kendall $\tau = 1.0$ for the top-two positions across all pairwise cohort
comparisons. The apparent stability is striking.

\begin{table*}[ht!]
\centering
\caption{Complete cross-cohort transfer. 7-feature ElasticNet trained on
  2022 medical students; applied without retraining. BDCI thresholds
  fixed at 2022 training-split percentiles.}
\label{tab:transfer}
\renewcommand{\arraystretch}{1.15}
\small
\begin{tabular}{@{}llccc@{}}
\toprule
\textbf{Cohort} & \textbf{Description} & $n$
  & \textbf{R\textsuperscript{2}} & \textbf{MAE} \\
\midrule
2022 Test (7-feat)         & Medical, reference split          & 178 & 0.423 & 9.23 \\
2022 Test (12-feat)        & Full feature model                & 178 & 0.414 & 9.26 \\
\midrule
Wave~1 (Mar.\ 2021)  & Longitudinal medical              & 898 & 0.443 & 8.89 \\
Wave~2 (Nov.\ 2021)  & Longitudinal medical              & 723 & 0.487 & 7.98 \\
Wave~3 (Nov.\ 2022)  & Longitudinal medical              & 959 & 0.388 & 8.67 \\
\midrule
NMS                        & Non-medical, 8 faculties         & 701 & 0.493 & 8.89 \\
\bottomrule
\end{tabular}
\end{table*}

Section~\ref{sec:audit} tests whether this stability reflects a portable
risk structure or a psychometric artefact.

\subsection{Residualization Audit: What Survives When the Depression Channel Is Removed?}
\label{sec:audit}

Table~\ref{tab:audit} is the central result. Two baseline comparisons
(primary BDCI and BDCI-A, an equal-weight variant) confirm that STAI-T
dominates at rank~1 regardless of the weighting scheme. When CES-D is
simply excluded from the outcome (BDCI-B), STAI-T remains rank~1 and
R\textsuperscript{2} falls from 0.414 to 0.303, a partial drop, because
the MBI-SS burnout subscales themselves correlate with depression.

The two residualization experiments go further. When STAI-T is replaced
by residuals orthogonal to CES-D (Experiment~1), R\textsuperscript{2}
drops from 0.414 to 0.156 and STAI-T falls from rank~1 to rank~6 in
mean absolute SHAP magnitude; health satisfaction becomes the dominant
predictor. When the burnout outcome itself is
purged of CES-D variance (Experiment~2, BDCI-C), R\textsuperscript{2}
collapses to 0.016: the full 12-feature model explains essentially none of
the burnout-specific variance.

\begin{table*}[ht!]
\centering
\caption{Residualization audit. ElasticNet fitted under five conditions
  using the same 80/20 split and hyperparameter grid. The central finding
  is the R\textsuperscript{2} collapse in the final two rows.}
\label{tab:audit}
\renewcommand{\arraystretch}{1.3}
\small
\begin{tabularx}{\textwidth}{@{}lXcc p{4.6cm}@{}}
\toprule
\textbf{Analysis} & \textbf{Outcome / predictor} & \textbf{R\textsuperscript{2}}
  & \textbf{STAI-T rank} & \textbf{Interpretation} \\
\midrule
Primary BDCI
  & STAI-T + MBI-SS + CES-D composite
  & 0.414 & 1 & Apparent strong signal \\
BDCI-A
  & Equal-weight composite (no amplification)
  & $\approx 0.42$ & 1 & Same pattern; robust to weighting \\
BDCI-B
  & MBI-SS only (CES-D excluded)
  & 0.303 & 1 & Burnout-related signal persists;
                MBI--CES-D correlation propagates \\
BDCI-C
  & MBI-SS residualized against CES-D
  & 0.016 & $\cdot$ & Burnout-specific variance is essentially
                       unpredictable from this feature set \\
STAI-T residualized
  & STAI-T $\perp$ CES-D as predictor
  & 0.156 & 6 & Anxiety--depression channel accounts for
                $\approx$60\% of model signal \\                       
\bottomrule
\end{tabularx}
\end{table*}

\begin{figure*}[ht!]
  \centering
  \includegraphics[width=\textwidth]{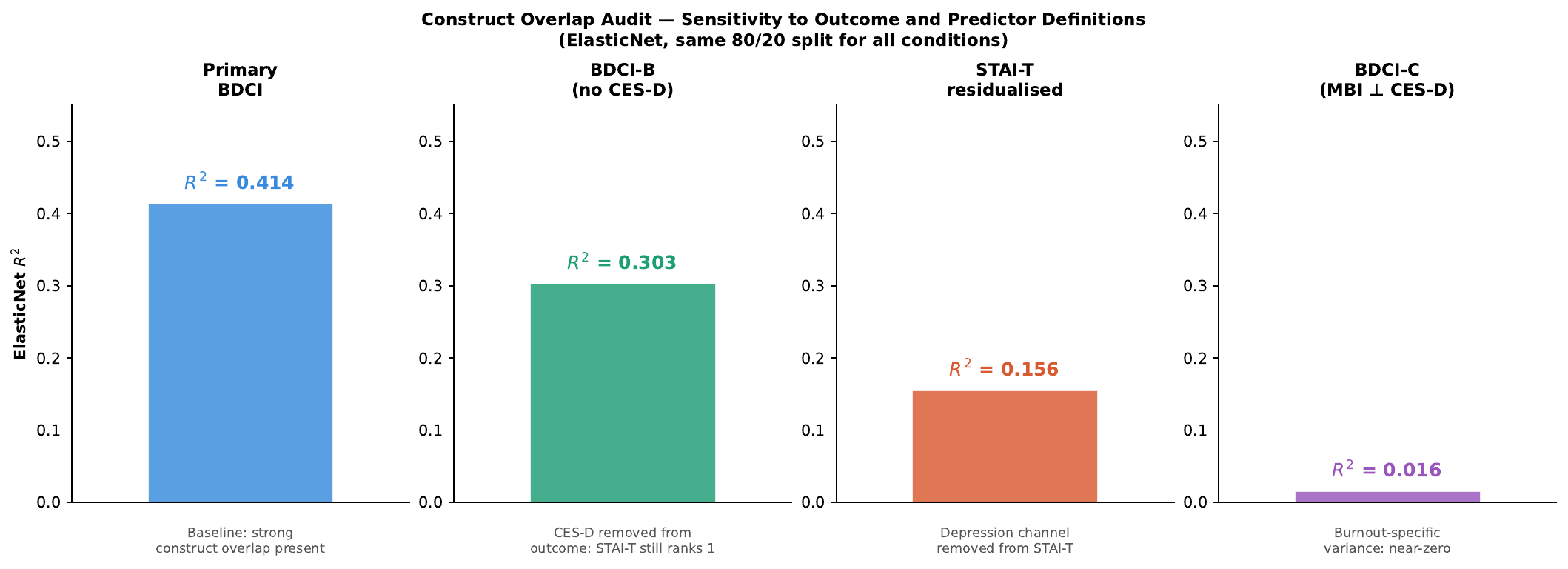}
  \caption{Construct-overlap residualization audit. ElasticNet
    R\textsuperscript{2} under four outcome/predictor definitions using the
    same 80/20 split. Performance drops from R\textsuperscript{2} = 0.414
    for the primary BDCI to 0.156 after residualizing STAI-T against
    CES-D, and to 0.016 when burnout subscales are residualized against
    CES-D (\textit{BDCI-C}). BDCI-B (CES-D excluded from outcome) yields
    R\textsuperscript{2} = 0.303, confirming that STAI-T's dominance
    persists even when the depression component of the outcome is removed.}
  \label{fig:audit}
\end{figure*}

Figure~\ref{fig:audit} summarises the residualization audit by plotting
ElasticNet R\textsuperscript{2} under each outcome/predictor definition.
The collapse from the primary BDCI to STAI-T residualization and BDCI-C
shows how much of the apparent signal depends on the anxiety--depression
channel.

These experiments answer the central question precisely. The standard
pipeline predicts the depression-saturated portion of the composite
outcome well and the burnout-specific portion essentially not at all.
The cross-cohort stability in Section~\ref{sec:transfer} transfers because
the STAI-T/CES-D psychometric correlation ($r = 0.72$) is structurally
constant wherever these instruments are used together, not because the
pipeline learned a portable multivariate risk structure.

Health satisfaction is the most robust predictor of the
depression-orthogonal signal (rank~1 in Experiment~1) and is the feature
most likely to represent an independent burnout pathway. Researchers who report cross-cohort XAI
stability without a residualization audit may be describing instrument
correlation, not replicated science.

\subsection{Uncertainty Rules Out Individual Deployment}
\label{sec:uncertainty}

\begin{figure*}[ht!]
  \centering
  \includegraphics[width=\textwidth]{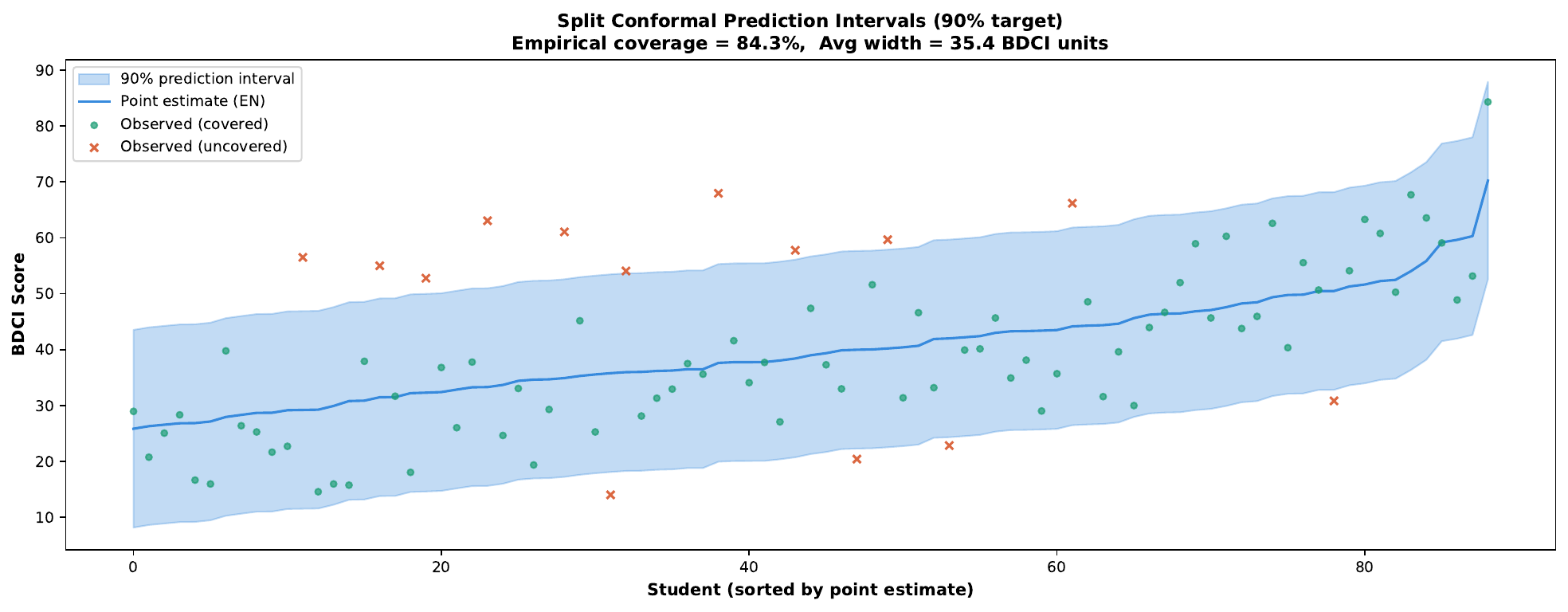}
  \caption{Split conformal prediction intervals (90\% target) for the
    ElasticNet model on the evaluation half ($n=89$), sorted by point
    estimate. Shaded region: prediction band of width $2\hat{q} = 35.4$
    BDCI units (2.4 outcome SDs). Cross markers: uncovered observations
    (miss rate: 15.7\%).}
  \label{fig:conformal}
\end{figure*}

The non-conformity quantile from the calibration half was $\hat{q} = 17.7$
BDCI units. On the evaluation half ($n_{\mathrm{eval}} = 89$), empirical
coverage was 84.3\% (nominal target: 90\%), and average interval width
was 35.4 BDCI units (Figure~\ref{fig:conformal}). The $-5.7$ percentage-point
coverage gap is within sampling noise: the standard deviation of a coverage
estimate at true 90\% with $n_{\mathrm{eval}} = 89$ is $\approx 3.2$ pp,
so the gap is approximately 1.8 standard deviations from nominal
($p \approx 0.07$). The headline finding is the interval width, not the
coverage deviation.

An interval of 35.4 units on a 0--100 scale is 2.4 outcome standard
deviations ($SD = 14.7$). This width is fully determined by the model's
residual distribution; any uncertainty quantification method would produce
comparable intervals. It independently establishes that individual-level
clinical prediction is not actionable with this feature set, regardless of
the construct overlap finding.

%% =============================================================================
\section{Discussion}
\label{sec:discussion}
%% =============================================================================

This study's central finding is the R\textsuperscript{2} collapse: from 0.414
(primary BDCI) to 0.016 (burnout-specific BDCI-C), with an intermediate
collapse to 0.156 when only the predictor is residualized. The implication
is not that the primary model is wrong: the anxiety--depression nexus is
one of the best-replicated constructs in psychiatric epidemiology, and the
STAI-T/CES-D association ($r = 0.72$) reflects genuine comorbidity
\citep{Barlow2014}. The implication is that a standard multi-method XAI
pipeline cannot distinguish this real clinical signal from a psychometric
artefact without an explicit orthogonality check. Apparent stability across
five cohorts is not evidence of a portable risk structure when the dominant
predictor and the dominant outcome component share a latent construct.

The residualization protocol introduced here is the paper's transferable
contribution. Any XAI study combining correlated predictor and outcome
constructs (trait anxiety and depression are a prominent example, but
any construct pair with $r > 0.5$--0.6 warrants scrutiny) should apply
this check before interpreting feature hierarchies as findings. The protocol
requires only the fitted model and a linear regression of the predictor on
the confounded outcome component; it adds no computational burden and
produces a clear, quantitative audit.

The apparently strong cross-cohort transfer (R\textsuperscript{2}: 0.41--0.49
across five evaluation sets) deserves an honest characterisation.
Wave~2 (0.487) and the non-medical cohort (0.493) exceed the internal
test R\textsuperscript{2} (0.423), but this is consistent with a stable
two-variable relationship rather than broad generalisation: all three
cohorts originate from the same institution and use identical instruments.
Cross-institutional validation with heterogeneous instruments (instruments
that break the STAI-T/CES-D fixed correlation) remains the essential next
step before any deployment.

The conformal width (35.4 BDCI units, 2.4 SDs) establishes a separate and
independent bar for deployment. Even if the construct overlap were resolved
by a better instrument design, individual-level prediction would require a
substantially expanded feature set (likely ecological momentary assessment,
physiological data, or longitudinal trajectories) to achieve actionable
precision. The current paper quantifies the upper bound of what cross-sectional
survey data can deliver under this instrument set.

Health satisfaction emerges from the residualization experiments as the most
durable predictor: after removing the STAI-T/CES-D channel, it ranks first
in mean absolute SHAP magnitude. This aligns with \citet{Bozorgmehr2023}
and \citet{Khan2024}, who identified health satisfaction and self-rated health
as primary determinants in unrelated datasets and populations, suggesting
that perceived health quality may be a more portable burnout signal than trait
anxiety. Testing this hypothesis requires a study that separates the
measurement constructs from the outset, not a post-hoc residualization.

\subsection{Limitations}
\label{sec:limits}

All datasets originate from the University of Lausanne; cross-institutional
generalisability has not been tested. The primary analysis is cross-sectional;
longitudinal within-student prediction would require a prospective design.
The BDCI is a study-internal composite that has not been independently
psychometrically validated. Potentially important predictors (financial
stress, sleep quality, social support, institutional climate) are absent
from the dataset and likely account for a substantial share of the
unexplained variance. The clinical-year calibration gap
(R\textsuperscript{2} = 0.294, $n = 66$, versus 0.456 for pre-clinical)
is directionally informative but statistically fragile at this subgroup size.

\subsection{Future Directions}
\label{sec:future}

The natural Phase~II is a multi-site prospective study that: (a) recruits
at least three institutions with structurally different curricula to obtain
genuine cross-institutional variation; (b) uses a prospective design in
which pre-year surveys predict end-of-year burnout episodes or counselling
referrals; (c) pre-registers the residualization audit as a primary
endpoint to confirm that the collapse is not an artefact of this particular
cohort; and (d) extends the clinical-year feature set with ecological
momentary assessment during clinical rotations to test the environmental
hypothesis.

%% =============================================================================
\section{Conclusion}
\label{sec:conclusion}
%% =============================================================================

A standard XAI pipeline applied to a burnout-depression composite produces
an apparently-robust, cross-population-stable risk hierarchy. Two
residualization experiments show that this hierarchy is largely an artefact
of construct overlap: removing the depression channel from the predictor
(STAI-T $\perp$ CES-D) drops R\textsuperscript{2} from 0.41 to 0.16;
removing it from the outcome (MBI-SS subscales $\perp$ CES-D) drops
R\textsuperscript{2} to 0.016. The apparent cross-cohort stability
(R\textsuperscript{2} = 0.41--0.49, Kendall $\tau = 1.0$ for top-two
positions across five cohorts) is explained by the fixed STAI-T/CES-D
psychometric correlation ($r = 0.72$), not by a learned portable risk
structure. Conformal prediction intervals averaging 35.4 BDCI units
(2.4 outcome SDs) independently establish that individual-level prediction
is not actionable.

The residualization protocol is directly transferable: any XAI study
that combines correlated predictor and outcome constructs should apply
this check before reporting apparent stability as a finding. Researchers
who skip this step may be describing instrument correlation, not replicated
science.

\section*{Data and Code Availability}
The primary 2022 cross-sectional dataset is available from the authors
upon reasonable request, subject to ethical approval. The temporal
longitudinal cohort is openly available under CC-BY~4.0 at Zenodo
(DOI: 10.5281/zenodo.8405764) \cite{Carrard2024}. The non-medical
student cohort is openly available under CC-BY~4.0 at Zenodo
(DOI: 10.5281/zenodo.15149289) \cite{Carrard2025}. Analysis code (Python)
is available at \url{https://github.com/alirezadehghan1/xai-burnout-depression}.

\section*{Conflict of Interest}
The authors declare no conflict of interest.

\section*{Ethical Approval}
The study uses anonymised secondary data collected under ethical approval
of the University of Lausanne Faculty of Biology and Medicine
(Protocol No.\ 2019-02134).

\bibliographystyle{unsrtnat}
\bibliography{references}

\end{document}